\documentclass[11pt,a4paper]{article}
\usepackage[hyperref]{emnlp-ijcnlp-2019}
\usepackage{times}
\usepackage{latexsym}
\usepackage{multirow}
\usepackage{multicol}
\usepackage{url}
\usepackage{graphicx}
\usepackage{subcaption}
\usepackage{amsfonts}
\usepackage[normalem]{ulem}
\aclfinalcopy 

\title{Local Contextual Attention with Hierarchical Structure for Dialogue Act Recognition}

\author{Zhigang Dai\textsuperscript{1}, Jinhua Fu\textsuperscript{2}, Qile Zhu\textsuperscript{3}, Hengbin Cui\textsuperscript{2}, Xiaolong Li\textsuperscript{2} and Yuan Qi\textsuperscript{2}\\ \textsuperscript{1}South China University of Technology \\
\textsuperscript{2} Ant Financial Service Group}

\date{}

\begin{document}
\maketitle
\begin{abstract}
Dialogue act recognition is a fundamental task for an intelligent dialogue system. Previous work models the whole dialog to predict dialog acts, which may bring the noise from unrelated sentences.
In this work, we design a hierarchical model based on self-attention to capture intra-sentence and inter-sentence information. 
We revise the attention distribution to focus on the local and contextual semantic information by incorporating the relative position information between utterances.
Based on the found that the length of dialog affects the performance, we introduce a new dialog segmentation mechanism to analyze the effect of dialog length and context padding length under online and offline settings. 
The experiment shows that our method achieves promising performance on two datasets: Switchboard Dialogue Act and DailyDialog with the accuracy of 80.34\% and 85.81\% respectively. Visualization of the attention weights shows that our method can learn the context dependency between utterances explicitly. 

\end{abstract}

\section{Introduction}

Dialogue act (DA) characterizes the type of a speaker's intention in the course of producing an utterance and is approximately equivalent to the illocutionary act of~\citet{Austin1962-AUSHTD} or the speech act of~\citet{searle1969speech}.
The recognition of DA is essential for modeling and automatically detecting discourse structure, especially in developing a human-machine dialogue system. 
It is natural to predict the \textsl{Answer} acts following an utterance of type \textsl{Question}, and then match the \textsl{Question} utterance to each QA-pair in the knowledge base. The predicted DA can also guide the response generation process~\cite{zhao2017learning}. For instance, system generates a \textsl{Greeting} type response to former \textsl{Greeting} type utterance. Moreover, DA is beneficial to other online dialogue strategies, such as conflict avoidance~\cite{nakanishi2018generating}.
In the offline system, DA also plays a significant role in summarizing and analyzing the collected utterances. For instance, recognizing DAs of a wholly online service record between customer and agent is beneficial to mine QA-pairs, which are selected and clustered then to expand the knowledge base. DA recognition is challenging due to the same utterance may have a different meaning in a different context. Table \ref{example} shows an example of some utterances together with their DAs from Switchboard dataset. In this example, utterance ``Okay.'' corresponds to two different DA labels within different semantic context.


\begin{table}
    \centering
    \small
    \begin{tabular}{l|l}
        \hline
         DA & Utterance \\
         \hline
         conventional-opening & B: Hi, \\
         conventional-opening & B: this is Donna Donahue.  \\
         conventional-opening & A: Hi, Donna.  \\
         conventional-opening & B: Hi.  \\
         yes-no-question & A: Ready to get started?  \\
         yes answers & B: Uh,  yeah,   \\
         statement-non-opinion & B: I think so.  \\
         other & A: Okay. \\
         \multirow{2}{*}{statement-non-opinion} & A: Sort of an interesting topic since \\
         & I just got back from lunch here. \\
         acknowledge & B: Okay. \\
         \hline
    \end{tabular}
    \caption{A snippet of a conversation with the DA labels from Switchboard dataset.}\label{example}
\end{table}

Many approaches have been proposed for DA recognition. Previous work relies heavily on handcrafted features which are domain-specific and difficult to scale up~\cite{stolcke2000dialogue,kim2010classifying,tavafi2013dialogue}. Recently, with great ability to do feature extraction, deep learning has yielded state-of-the-art results for many NLP tasks, and also makes impressive advances in DA recognition.
~\citet{liu2017using,bothe2018context} built hierarchical CNN/RNN models to encode sentence and incorporate context information for DA recognition.
~\citet{kumar2018dialogue} achieved promising performance by adding the CRF to enhance the dependency between labels.
~\citet{raheja2019dialogue} applied the self-attention mechanism coupled with a hierarchical recurrent neural network.

However, previous approaches cannot make full use of the relative position relationship between utterances. It is natural that utterances in the local context always have strong dependencies in our daily dialog.
In this paper, we propose a hierarchical model based on self-attention~\cite{vaswani2017attention} and revise the attention distribution to focus on a local and contextual semantic information by a learnable Gaussian bias which represents the relative position information between utterances, inspired by \citet{yang2018modeling}.
Further, to analyze the effect of dialog length quantitatively, we introduce a new dialog segmentation mechanism for the DA task and evaluate the performance of different dialogue length and context padding length under online and offline settings.
Experiment and visualization show that our method can learn the local contextual dependency between utterances explicitly and achieve promising performance in two well-known datasets.

The contributions of this paper are:
\begin{itemize}
\item We design a hierarchical model based on self-attention and revise the attention distribution to focus on a local and contextual semantic information by the relative position information between utterances.  
\item We introduce a new dialog segmentation mechaism for the DA task and analyze the effect of dialog length and context padding length.
\item In addition to traditional offline prediction, we also analyze the accuracy and time complexity under the online setting.
\end{itemize}

\section{Background}
\subsection{Related Work} \label{DA_Recognition}
DA recognition is aimed to assign a label to each utterance in a conversation. It can be formulated as a supervised classification problem. There are two trends to solve this problem: 1) as a sequence labeling problem, it will predict the labels for all utterances in the whole dialogue history~\cite{dielmann2008recognition,lee2016sequential,kumar2018dialogue}; 2) as a sentence classification problem, it will treat utterance independently without any context history~\cite{kim2010classifying,khanpour2016dialogue}. Early studies rely heavily on handcrafted features such as lexical, syntactic, contextual, prosodic and speaker information and achieve good results~\cite{dielmann2008recognition,stolcke2000dialogue,chen2013multimodality}.

Recent studies have applied deep learning based model for DA recognition.
~\citet{lee2016sequential} proposed a model based on RNNs and CNNs that incorporates preceding short texts to classify current DAs.
~\citet{liu2017using,bothe2018context} used hierarchical CNN and RNN to model the utterance sequence in the conversation, which can extract high-level sentence information to predict its label. They found that there is a small performance difference among different hierarchical CNN and RNN approaches.
~\citet{kumar2018dialogue} added a CRF layer on the top of the hierarchical network to model the label transition dependency. ~\citet{raheja2019dialogue} applied the context-aware self-attention mechanism coupled with a hierarchical recurrent neural network and got a significant improvement over state-of-the-art results on SwDA datasets.
On another aspect,~\citet{ji2016latent} combined a recurrent neural network language model with a latent variable model over DAs.
~\citet{zhao2018dg} proposed a Discrete Information Variational Autoencoders (DI-VAE) model to learn discrete latent actions to incorporate sentence-level distributional semantics for dialogue generation.
\begin{figure*}[h!]
  \includegraphics[width=\linewidth]{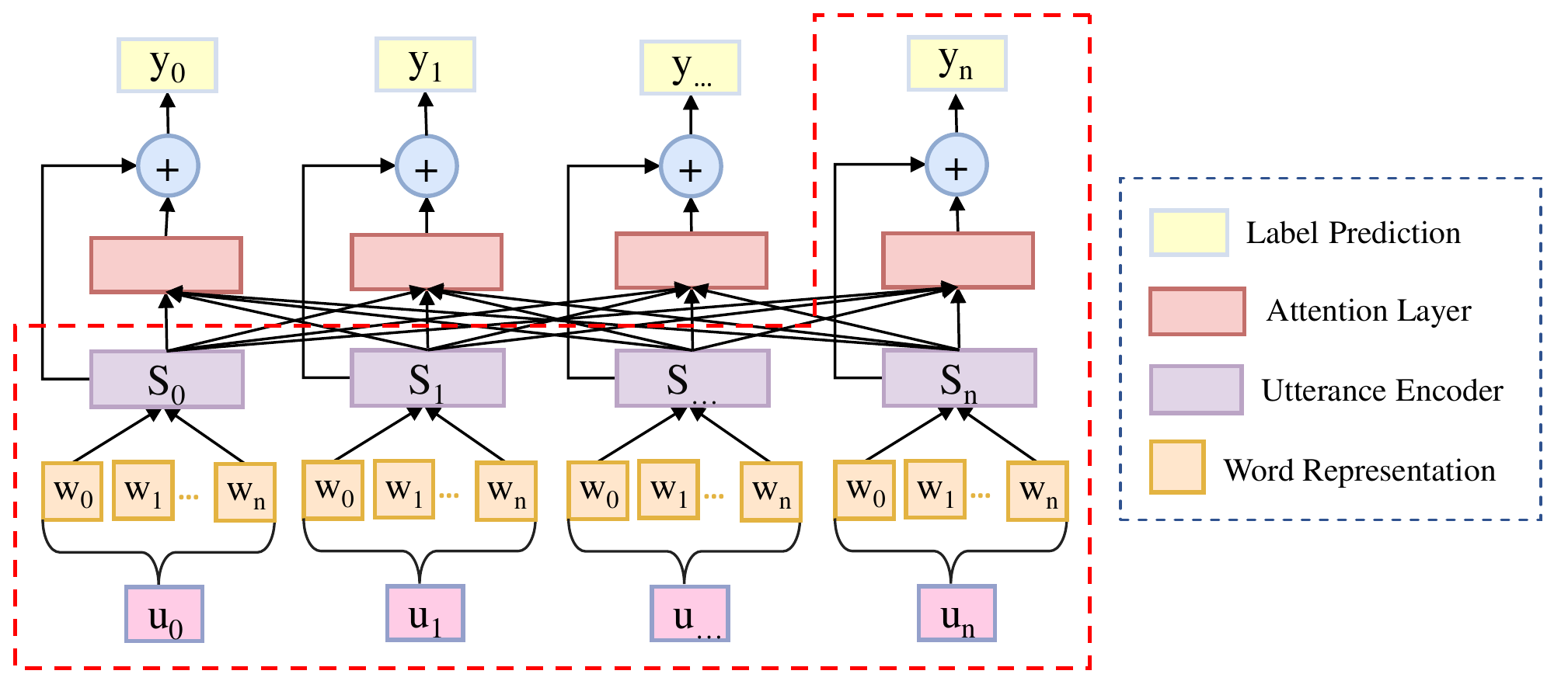}
  \caption{The model structure for DA recognition, where the LSTM with max pooling is simplified as utterance encoder in our experiment. The area in the red dashed line represents the structure for online prediction.}
  \label{context_attention}
\end{figure*}
\subsection{Self-Attention} \label{Self-Attention}
Self-attention~\cite{vaswani2017attention} achieves great success for its efficiently parallel computation and long-range dependency modeling. 

Given the input sequence $ s = \left( s_1,...,s_n \right) $ of n elements where $ s_i \in \mathbb{R}^{d_s} $. 
Each attention head holds three parameter matrices, $W_h^Q, W_h^K, W_h^V \in {\mathbb{R}}^{d_s \times d_z} $ where $ h $ present the index of head. For the head $h$, linear projection is applied to the sequence $s$ to obtain key (K), query (Q), and value (V) representations. the attention module gets the weight by computing dot-products between key/query pair and then $softmax$ normalizes the result. it is defined as:
$$
ATT_h(Q,K,V) = softmax(\frac{QK^{T}}{\sqrt{d_z}})\times V,
$$
where $\sqrt{d_z}$ is the scaling factor to counteract this effect that the dot products may grow large in magnitude.
For all the heads, 
$$
output = Concat(ATT_1,...,ATT_h)\times W^O,
$$
where $W^O \in \mathbb{R}^{(d_z*h)\times d_s}$ is the output projection.

One weakness of self-attention model it that they cannot encode the position information efficiently. Some methods have been proposed to encode the relative or absolute position of tokens in the sequence as the additional input to the model.~\citet{vaswani2017attention} used sine and cosine functions of different frequencies and added positional encodings to the input embeddings together. It used absolute position embedding to capture relative positional relation by the characteristic of sine and cosine functions. 
Moreover, several studies show that explicitly modeling relative position can further improve performance. For example,~\citet{shaw2018self} proposed relative position encoding to explicitly model relative position by independent semantic parameter. 
It demonstrated significant improvements even when entirely replacing conventional absolute position encodings.~\citet{yang2018modeling} proposed to model localness for the self-attention network by a learnable Gaussian bias which enhanced the ability to model local relationship and demonstrated the effectiveness on the translation task. 

In our study, we design a local contextual attention model, which incorporates relative position information by a learnable Gaussian bias into original attention distribution. 
Different from~\citet{yang2018modeling}, in our method, the distribution center is regulated around the corresponding utterance with a window, which indicates the context dependency preference, for capturing more local contextual dependency. 

\section{Methodology} \label{Methodology}
Before we describe the proposed model in detail, we first define the mathematical notation for the DA recognition task in this paper. Given the dataset, $X = (D_1,D_2,... D_L)$ with corresponding DA labels $(Y_1,Y_2,...Y_L)$. Each dialogue is a sequence of $ N_l $ utterances $ D_l = (u_1,u_2,...u_{N_l})$ with $ Y_l = (y_1,y_2,...y_{N_l}) $. Each utterance is padded or truncated to the length of $ M $ words, $u_j = (w_1,w_2,...w_{M})$.

Figure~\ref{context_attention} shows our overall model structure. For the first layer, we encode each utterance $u_j$ into a vector representation. Each word $w_m$ of the utterance $u_j$ is converted into dense vector representations $e_m$ from one-hot token representation. And then, we apply LSTM~\cite{hochreiter1997long}, a powerful and effective structure for sequence modeling, to encode the word sequence. Formally, for the utterance $u_j$:
\begin{eqnarray}
	e_m = embed(w_m) \quad \forall m\in \{1,...M\}, \\
	h_m = LSTM(h_{m-1},e_m)  \quad \forall m\in \{1,...M\},
\end{eqnarray}
where $embed$ represents the embedding layer which can be initialized by pre-trained embeddings. To make a fair comparison with previous work, we do not use the fine-grained embedding presented in~\citet{chen2018dialogue}. LSTM helps us get the context-aware sentence representation for the input sequence. There are several approaches to represent the sentence from the words. Following~\citet{Alexis2017Supervised}, we add a max-pooling layer after LSTM, which selects the maximum value in each dimension from the hidden units. In our experiment, LSTM with max-pooling does perform a little better than LSTM with last-pooling, which is used in ~\citet{kumar2018dialogue}.

Afterwards, we get the utterances vector representations $ u = (u_1,...,u_{N_l}) $ of $N_l$ elements for the dialogue $D_l$ where $ u_j \in \mathbb{R}^{d_s}, d_s$ is the dimension of hidden units.
As we discussed in section~\ref{Self-Attention}, given the sequence $ s \in \mathbb{R}^{N_l*d_s}$, self-attention mechanism calculates the attention weights between each pair of utterances in the sequence and get the weighted sum as output. The attention module explicitly models the context dependency between utterances. We employ a residual connection~\cite{he2016deep} around the attention module, which represents the dependency encoder between utterances, and the current utterance encoder $s$: 
\begin{equation}
output = output + s.
\end{equation}


Finally, we apply a two-layer fully connected network with a Rectified Linear Unit (ReLU) to get the final classification output for each utterance.


\subsection{Modeling Local Contextual Attention}
The attention explicitly models the interaction between the utterances. However, for context modeling, original attention mechanism always considers all of the utterances in a dialogue which inhibits the relation among the local context and is prone to overfitting during training. It is natural that utterances in the local context always have strong dependencies in our daily dialog. Therefore, we add a learnable Gaussian bias with the local constraint to the weight normalized by $softmax$ to enhance the interaction between concerned utterances and its neighbors.

The attention module formula is revised as:
\begin{equation}
ATT(Q,K) = softmax(\frac{QK^{T}}{\sqrt{d}} + POS).
\end{equation}
The first term is the original dot product self-attention model. $POS \in \mathbb{R}^{N\times N}$ is the bias matrix, where N is the length of dialogue. 
The element $POS_{i,j}$ is defined following by gaussian distribution:
\begin{equation}
POS_{i,j} = -\frac{(j-c_{i})^2}{2w_{i}^2},
\end{equation}
$POS_{i,j}$ measures the dependency between the utterance $u_j$ and the utterance $u_i$ in terms of the relative position prior.
$w_{i}$ represents for the standard deviation, which controls the weight decaying.
Because of local constraint, $|c_{i} - i| <= C$, for each utterance $u_i$, the predicted center position $c_{i}$ and window size $ w_{i}$ is defined as followed: 
\begin{eqnarray}
	c_{i} = i + C\times tanh(W_i^c\times \overline K), \\
	w_i = D\times sigmoid(W_i^d\times \overline K),
\end{eqnarray}
where $W_i^c,W_i^d \in \mathbb{R}^{1*N}$ are both learnable parameters. We initialized the parameter $W_i^c$ to 0, which leads to center position $ c_i = i $ by default. 
Furthermore, $c_{i}$ and $w_{i}$ are both related to the semantic context of the utterances, so we assign the mean of key $\overline K$ in attention mechanism to represent the context information. Moreover, the central position also indicates the dependency preference of the preceding utterances or subsequent utterances. 

It is worth noting that there is a little difference with ~\citet{yang2018modeling}, although we both revise the attention module by the Gaussian distribution. In our method, for the given utterance $u_{i}$, the distribution center $c_{i}$ is regulated for capturing the not only local but also contextual dependency, which can be formally expressed as: $c_{i} \in (i-C,i+C)$. However, in their work, the distribution center can be anywhere in the sequence, and it is designed for capturing the phrasal patterns, which are essential for Neural Machine Translation task.



\subsection{Online and Offline Predictions}
Previous work mainly focuses on the offline setting where we can access the whole utterances in the dialogue and predict all the DA labels simultaneously. However, the online setting is the natural demand in our real-time applications. For the online setting, we only care about the recognition result of the last utterance in the given context, as seen in the area with the red dashed line in Figure~\ref{context_attention}, our model is well compatible with online setting, we can calculate the attention between the last utterance and the other utterances directly where $K \in \mathbb{R}^{1\times d}, Q \in \mathbb{R}^{n\times d}, V \in \mathbb{R}^{n\times d}$. For LSTM, we still have to model the entire sequence, which is slower than attention based models. Table~\ref{time complexity} shows the time complexity comparison excluding the time cost of first layer encoding, and the dialogue length $n$ is smaller than the representation dimension $d$. Our model is easy to expand into the online setting, however, to have a fair comparison with previous work, in our experiments, we applied the models under the offline setting by default.

\subsection{Separate into Sub-dialogues}
The length of different dialogues in the dataset varies a lot. It is worth noting that the length of dialog affects the model prediction. On the one hand, under the offline setting, we can access the whole utterances in the dialogue and predict all the DA labels simultaneously, so the more utterances, the more efficient. However, on the other hand, if we put too many utterances in once prediction, it will model too much unrelated dependency in the long utterances sequence for both LSTM and attention mechanism based model. The sub-dialogues with the same length also enable efficiently batch training. To study how the dialogue length and context padding length will affect the performance, so we defined a sliding window $W$ which is the sub-dialogue length. Then, we separate each long dialogue into several small sub-dialogues. For example, the dialog $D$ is a sequence of utterances with length $n$, and we will get $\lceil x/w \rceil $ sub-dialogues, for the k-th sub-dialogues, the utterances sequence is $(u_{(k-1)*W+1},u_{(k-1)*W+2},...,u_{k*W})$.
In order to avoid losing some context information caused by being separated, which will affect the context modeling for the utterances in the begin and end of the sub-dialog, we add the corresponding context with $P$ (stands for context padding) utterances at the begin and the end of each sliding window, so for the k-th sub-dialogues, the revised utterances sequence is $(u_{(k-1)*W-P+1},u_{(k-1)*W-P+2},...,u_{k*W+P})$. Moreover, we mask the loss for the context padding utterances, which can be formally expressed as:

\begin{equation}
loss = \frac{1}{W} \sum_{i}M(i) L(\hat{y_i},y_i), 
\end{equation}
$M(i)=0$ if utterance $i$ is in the context padding otherwise 1, $L$ is the cross entropy.

The $W$ and $P$ are both hyperparameters; in the experiment \ref{swda_result}, we will talk about the effect of the window size and the context padding length.


\begin{table}
    \centering
    \begin{tabular}{l|l|l}
        \hline
         model & offline setting & online setting \\
         \hline
         LSTM & $n\times d^2$ & $n\times d^2$ \\
         \hline
         Self-Attention & $ n^2\times d $ & $n\times d$ \\
         \hline
    \end{tabular}
    \caption{Time complexity between LSTM and self-attention for both online and offline predictions excluding the time cost of first layer encoding. The parameter n represents for the dialogue length in the sliding window and d represent for the dimension of representation unit. }
    \label{time complexity}
\end{table}

\section{Experiments}

\begin{table}
\centering
\small

\begin{tabular}{l|l|l|l|l|l}
\hline  
Dataset & $|C|$ & $|U|$ & train & validation & test \\
\hline  
SwDA & 42 & 176 & 1K(177K) & 112(18K) & 19(4K) \\
Daily & 4 & 8 & 11K(87K) & 1K(8K) & 1K(8K) \\
\hline 
\end{tabular}
\caption{\label{statistics}$|C|$ indicates the number of classes. $|U|$ indicates the average length of dialogues. The train/validation/test columns indicate the number of dialogues (the number of sentences) in the respective splits.}

\end{table}

\subsection{Datasets}
We evaluate the performance of our model on two high-quality datasets: Switchboard Dialogue Act Corpus (SwDA)~\cite{stolcke2000dialogue} and DailyDialog~\cite{I17-1099}. SwDA has been widely used in previous work for the DA recognition task. It is annotated on 1155 human to human telephonic conversations about the given topic. Each utterance in the conversation is manually labeled as one of 42 dialogue acts according to SWBD-DAMSL taxonomy~\cite{jurafsky1997switchboard}. In \citet{raheja2019dialogue}, they used 43 categories of dialogue acts, which is different from us and previous work. The difference in the number of labels is mainly due to the special label ``+'', which represents that the utterance is interrupted by the other speaker (and thus split into two or more parts). We used the same processing with \citet{ milajevs2014investigating}, which concatenated the parts of an interrupted utterance together, giving the result the tag of the first part and putting it in its place in the conversation sequence. It is critical for fair comparison because there are nearly 8\% data has the label ``+''. Lacking standard splits, we followed the training/validation/test splits by~\citet{lee2016sequential}.
DailyDialog dataset contains 13118 multi-turn dialogues, which mainly reflect our daily communication style. It covers various topics about our daily life. Each utterance in the conversation is manually labeled as one out of 4 dialogue act classes. Table~\ref{statistics} presents the statistics for both datasets. In our preprocessing, the text was lowercased before tokenized, and then sentences were tokenized by WordPiece tokenizer~\cite{wu2016google} with a 30,000 token vocabulary to alleviate the Out-of-Vocabulary problem.

\begin{table}
\centering
\begin{tabular}{l|l}
\hline 
models & Acc(\%) \\
\hline
\hline
\multicolumn{2}{c}{previous approaches} \\
\hline
BLSTM+Attention+BLSTM~\shortcite{raheja2019dialogue}& 82.9  \\
Hierarchical BLSTM-CRF~\shortcite{kumar2018dialogue}& 79.2  \\
CRF-ASN \shortcite{chen2018dialogue}& 78.7\footnotemark[1]{} \\ 
Hierarchical CNN (window 4) \shortcite{liu2017using} & 78.3\footnotemark[2]{} \\
mLSTM-RNN ~\shortcite{bothe2018context} & 77.3 \\ 
DRLM-Conditional~\shortcite{ji2016latent} & 77.0 \\
LSTM-Softmax~\shortcite{khanpour2016dialogue} & 75.8 \\
RCNN~\shortcite{kalchbrenner2013recurrent} & 73.9 \\
CNN~\shortcite{lee2016sequential} & 73.1 \\
CRF~\shortcite{kim2010classifying} & 72.2 \\
\hline  
\hline
\multicolumn{2}{c}{reimplemented and proposed approaches} \\
\hline
CNN & 75.27 \\ 
LSTM & 75.59 \\
BERT\shortcite{devlin2018bert} & 76.88 \\
\hline
LSTM+BLSTM & 80.00 \\
LSTM+Attention & 80.12 \\
LSTM+Local Contextual Attention & \textbf{80.34} \\
\hline
Human annotator & 84.0 \\
\hline
\end{tabular}

\caption{\label{experiments}Comparison results with the previous approaches and our approaches on SwDA dataset.}

\end{table}
\footnotetext[1]{The author claimed that they achieved 78.7\%(81.3\%) accuracy with pre-trained word embedding (fine-grained embedding). For a fair comparison, both previous and our work is simply based on pre-trained word embedding.}
\footnotetext[2]{The author randomly selected two test sets which are different from previous and our work and achieved 77.15\% and 79.74\%, and we reimplemented in standard test sets.}

\subsection{Results on SwDA} \label{swda_result}

\begin{table}[t]
\centering
\begin{tabular}{c|l|l|l}
\hline 
$W$ & $P$ & models & Acc(\%) \\
\hline  
\multirow{3}{*}{1} & \multirow{3}{*}{0} & CNN & 75.27  \\
& & LSTM & 75.59 \\
& & BERT & 76.88 \\
\hline
\multirow{2}{*}{1} & \multirow{2}{*}{1} & LSTM+BLSTM & 78.60 \\
& &LSTM+Attention & 78.74 \\
\hline 
\multirow{2}{*}{1} & \multirow{2}{*}{3} & LSTM+BLSTM & 79.36 \\
& &LSTM+Attention & 79.98 \\
\hline 
\multirow{2}{*}{1} & \multirow{2}{*}{5} & \textbf{LSTM+BLSTM} & \textbf{80.00} \\
& & \textbf{LSTM+Attention} & \textbf{80.12} \\
\hline 
\multirow{3}{*}{5} & \multirow{3}{*}{5} & LSTM+BLSTM & 78.50 \\
& &LSTM+Attention & 79.43 \\
& &LSTM+LC Attention & 80.27 \\
\hline 
\multirow{3}{*}{10} & \multirow{3}{*}{5} & LSTM+BLSTM & 78.31 \\
& &LSTM+Attention & 79.00 \\
& &\textbf{LSTM+LC Attention} & \textbf{80.34} \\
\hline 
\multirow{3}{*}{20} & \multirow{3}{*}{5} & LSTM+BLSTM & 78.55 \\
& & LSTM+Attention & 78.57 \\
& & LSTM+LC Attention & 80.17 \\
\hline 
\hline 
\multicolumn{4}{c}{online prediction} \\
\hline 
\multicolumn{3}{c}{LSTM+LSTM} & 78.62 \\
\multicolumn{3}{c}{LSTM+Attention} & 78.86 \\
\multicolumn{3}{c}{LSTM+LC Attention} & \textbf{78.93} \\
\hline
\end{tabular}
\caption{\label{other_experiment}Experiment results about the hyperparameter $W$ and $P$ on SwDA dataset and online prediction result. $W,P$ indicate the size of sliding window and context padding length during training and testing.}
\end{table}
In this section, we evaluate the proposed approaches on SwDA dataset. Table~\ref{experiments} shows our experimental results and the previous ones on SwDA dataset. It is worth noting that \citet{raheja2019dialogue} combined GloVe\cite{pennington2014glove} and pre-trained ELMo representations\cite{peters2018deep} as word embeddings. However, in our work, we only applied the pre-trained word embedding. To illustrate the importance of context information, we also evaluate several sentence classification methods (CNN, LSTM, BERT) as baselines. For baseline models, both CNN and LSTM, got similar accuracy (75.27\% and 75.59\% respectively). We also fine-tuned BERT~\cite{devlin2018bert} to do recognition based on single utterance. As seen, with the powerful unsupervised pre-trained language model, BERT (76.88\% accuracy) outperformed LSTM and CNN models for single sentence classification. However, it was still much lower than the models based on context information. It indicates that context information is crucial in the DA recognition task. BERT can boost performance in a large margin. However, it costs too much time and resources. In this reason, we chose LSTM as our utterance encoder in further experiment.

By modeling context information, the performance of the hierarchical model is improved by at least 3\%, even compared to BERT.
In order to better analyze the semantic dependency learned by attention, in our experiments, we removed the CRF module.
In terms of different hierarchical models, our LSTM+BLSTM achieved good result. The accuracy was 80.00\% which is even a little better than Hierarchical BLSTM-CRF~\cite{kumar2018dialogue}. Relying on attention mechanism and local contextual modeling, our model, LSTM+Attention and LSTM+Local Contextual Attention, achieved 80.12\% and 80.34\% accuracy respectively. Compared with the previous best approach Hierarchical BLSTM-CRF, we can obtain a relative accuracy gain with 1.1\% by our best model. It indicated that self-attention model can capture context dependency better than the BLSTM model. With adding the local constraint, we can get an even better result.

To further illustrate the effect of the context length, we also performed experiments with different sliding window $W$ and context padding $P$. Table~\ref{other_experiment} shows the result. It is worth noting that it is actually the same as single sentence classification when $P = 0$ (without any context provided). 
First, we set $W$ to 1 to discuss how the length of context padding will affect. As seen in the result, the accuracy increased when more context padding was used for both LSTM+BLSTM and LSTM+Attention approaches, so we did not evaluate the performance of LSTM+LC Attention when context padding is small. 
There was no further accuracy improvement when the length of context padding was beyond 5. Therefore, we fixed the context padding length $P$ to 5 and increased the size of the sliding window to see how it works. With sliding window size increasing, the more context was involved together with more unnecessary information. From the experiments, we can see that both LSTM+BLSTM and LSTM+Attention achieved the best performance when window size was 1 and context padding length was 5. When window size increased, the performances of these two models dropped. However, our model (LSTM+LC Attention) can leverage the context information more efficiently, which achieved the best performance when window size was 10, and the model was more stable and robust to the different setting of window size.

For online prediction, we only care about the recognition result of the last utterance in the given context. We added 5 preceding utterances as context padding for every predicted utterance because we cannot access subsequent utterances in the online setting. As seen in Table~\ref{other_experiment}, without subsequent utterances, the performances of these three models dropped. However, LSTM+LC Attention still outperformed the other two models.

\begin{table}
\centering
\begin{tabular}{c|l|l|l}
\hline 
$W$ & $P$ & models & Acc(\%) \\
\hline  
\multirow{3}{*}{1} & \multirow{3}{*}{0} & CNN & 82.22  \\
& & LSTM & 82.58 \\
& & BERT & 83.22 \\
\hline
\multirow{2}{*}{1} & \multirow{2}{*}{1} & LSTM+BLSTM & 84.88 \\
& &LSTM+Attention & 85.10 \\
\hline 
\multirow{2}{*}{1} & \multirow{2}{*}{2} & LSTM+BLSTM & \textbf{85.06} \\
& &LSTM+Attention & \textbf{85.36} \\
\hline 
\multirow{2}{*}{1} & \multirow{2}{*}{3} & LSTM+BLSTM & 84.97 \\
& &LSTM+Attention & 85.05 \\
\hline 
\multirow{3}{*}{5} & \multirow{3}{*}{2} & LSTM+BLSTM & 85.01 \\
& &LSTM+Attention & 85.26 \\
& &LSTM+LC Attention & \textbf{85.81} \\
\hline 
\multirow{3}{*}{10} & \multirow{3}{*}{2} & LSTM+BLSTM & 84.97 \\
& &LSTM+Attention & 85.13 \\
& &\textbf{LSTM+LC Attention} & 85.72 \\
\hline 
\hline 
\multicolumn{4}{c}{online prediction} \\
\hline 
\multicolumn{3}{c}{LSTM+LSTM} & 84.55 \\
\multicolumn{3}{c}{LSTM+Attention} & 84.68 \\
\multicolumn{3}{c}{LSTM+LC Attention} & \textbf{84.83} \\
\hline
\end{tabular}
\caption{\label{dailydialog_experiment}Experiment results on DailyDialog dataset.}
\end{table}
\begin{figure*}[h!]
  \centering
  \begin{subfigure}[b]{0.48\linewidth}
    \includegraphics[width=\linewidth]{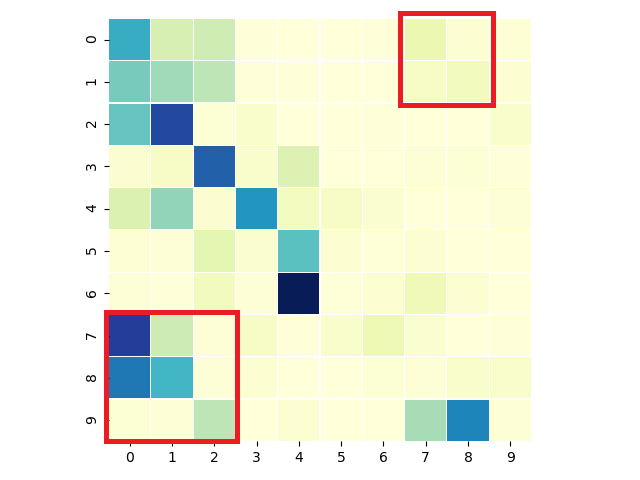}
    \caption{original attention weight matrix}
    \label{original_attention}
  \end{subfigure}
  \begin{subfigure}[b]{0.48\linewidth}
    \includegraphics[width=\linewidth]{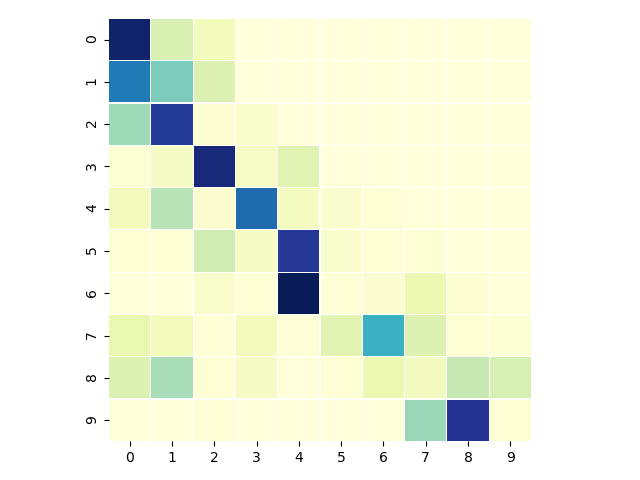}
    \caption{local contextual attention weight matrix}
    \label{gaussian_attention}
  \end{subfigure}
  \caption{Visualization of original attention and local contextual attention. Each colored grid represents the dependency score between two sentences. The deeper the color is, the higher the dependency score is.}
  \label{visualization}
\end{figure*}
\subsection{Result on DailyDialog}
The classification accuracy of DailyDialog dataset is summarized in Table \ref{dailydialog_experiment}. As for sentence classification without context information, the fine-tuned BERT still outperformed LSTM and CNN based models. From table ~\ref{statistics} we can see that, the average dialogue length $|U|$ in DailyDialog is much shorter than the average length of SwDA. So, in our experiment, we set the maximum of the $W$ to 10, which almost covers the whole utterances in the dialogue. Using the same way as SwDA dataset, we, first, set W to 1 and increased the length of context padding. As seen, modeling local context information, hierarchical models yielded significant improvement than sentence classification. There was no further accuracy improvement when the length of context padding was beyond 2, so we fixed the context padding length P to 2 and increased the size of sliding window size W. From the experiments, we can see that LSTM+Attention always got a little better accuracy than LSTM+BLSTM. With window size increasing, the performances of these two models dropped. Relying on modeling local contextual information, LSTM+LC Attention achieved the best accuracy (85.81\%) when the window size was 5. For the longer sliding window, the performance of LSTM+LC Attention was still better and more robust than the other two models. For online prediction, we added 2 preceding utterances as context padding, and the experiment shows that LSTM+LC Attention outperformed the other two models under the online setting, although the performances of these three models dropped without subsequent utterances.


\subsection{Visualization}
In this section, we visualize the attention weights for analyzing how local contextual attention works in detail. Figure~\ref{visualization} shows the visualization of original attention and local contextual attention for the example dialogue shown in Table~\ref{example}. The attention matrix $M$ explicitly measures the dependency among utterances. Each row of grids is normalized by $softmax$, $M_{ij}$ represents for the dependency score between the utterance i and utterance j.
As demonstrated in Figure~\ref{original_attention}, there are some wrong and uninterpretable attention weights annotated with red color, which is learned by the original attention. The original attention model gives the utterance ``B: Hi'' (position 0) and ``A: Okay.'' (position 7) a high dependency score. However, local contextual attention weakens its attention weights due to the long distance apart.

Overall, the additional Gaussian bias trend to centralize the attention distribution to the diagonal of the matrix, which is in line with our linguistic intuition that utterances that are far apart usually don't have too strong dependencies. As demonstrated in Figure~\ref{gaussian_attention}, benefiting of the additional Gaussian bias, the revised attention mechanism weakens the attention weights between utterances which cross the long relative distance. For the grids near diagonal, it strengthens their dependency score and doesn't bring other useless dependencies for its learnable magnitude.




\section{Conclusions and Future Work}
In the paper, we propose our hierarchical model with local contextual attention to the Dialogue Act Recognition task. Our model can explicitly capture the semantic dependencies between utterances inside the dialogue. 
To enhance our model with local contextual information, we revise the attention distribution by a learnable Gaussian bias to make it focus on the local neighbors. 
Based on our dialog segmentation mechanism, we find that local contextual attention reduces the noises through relative position information, which is essential for dialogue act recognition. And this segmentation mechanism can be applied under online and offline settings. Our model achieves promising performance in two well-known datasets, which shows that modeling local contextual information is crucial for dialogue act recognition.


There is a close relation between dialogue act recognition and discourse parsing~\cite{asher2016discourse}. 
The most discourse parsing process is composed of two stages: structure construction and dependency labeling~\cite{wang2017dp,shi2018dp}.
For future work, a promising direction is to apply our method to multi-task training with two stages jointly. Incorporating supervised information from dependency between utterances may enhance the self-attention and further improve the accuracy of dialogue act recognition.



\bibliography{emnlp-ijcnlp-2019}

\begin{thebibliography}{35}
\expandafter\ifx\csname natexlab\endcsname\relax\def\natexlab#1{#1}\fi

\bibitem[{Alexis et~al.(2017)Alexis, Douwe, Holger, Lo~̈ıc, and
  Antoine}]{Alexis2017Supervised}
Conneau Alexis, Kiela Douwe, Schwenk Holger, Barraul Lo~̈ıc, and Bordes
  Antoine. 2017.
\newblock Supervised learning of universal sentence representations from
  natural language inference data.
\newblock In \emph{Proceedings of the 2017 Conference on Empirical Methods in
  Natural Language Processing}, page 670–680.

\bibitem[{Asher et~al.(2016)Asher, Hunter, Morey, Benamara, and
  Afantenos}]{asher2016discourse}
Nicholas Asher, Julie Hunter, Mathieu Morey, Farah Benamara, and Stergos~D
  Afantenos. 2016.
\newblock Discourse structure and dialogue acts in multiparty dialogue: the
  stac corpus.
\newblock In \emph{LREC}.

\bibitem[{Austin(1962)}]{Austin1962-AUSHTD}
J.~L. Austin. 1962.
\newblock \emph{How to Do Things with Words}.
\newblock Clarendon Press.

\bibitem[{Bothe et~al.(2018)Bothe, Weber, Magg, and Wermter}]{bothe2018context}
Chandrakant Bothe, Cornelius Weber, Sven Magg, and Stefan Wermter. 2018.
\newblock A context-based approach for dialogue act recognition using simple
  recurrent neural networks.
\newblock In \emph{Proceedings of the Eleventh International Conference on
  Language Resources and Evaluation (LREC-2018)}.

\bibitem[{Chen and Di~Eugenio(2013)}]{chen2013multimodality}
Lin Chen and Barbara Di~Eugenio. 2013.
\newblock Multimodality and dialogue act classification in the robohelper
  project.
\newblock In \emph{Proceedings of the SIGDIAL 2013 Conference}, pages 183--192.

\bibitem[{Chen et~al.(2018)Chen, Yang, Zhao, Cai, and He}]{chen2018dialogue}
Zheqian Chen, Rongqin Yang, Zhou Zhao, Deng Cai, and Xiaofei He. 2018.
\newblock Dialogue act recognition via crf-attentive structured network.
\newblock In \emph{The 41st International ACM SIGIR Conference on Research \&
  Development in Information Retrieval}, pages 225--234. ACM.

\bibitem[{Devlin et~al.(2018)Devlin, Chang, Lee, and
  Toutanova}]{devlin2018bert}
Jacob Devlin, Ming-Wei Chang, Kenton Lee, and Kristina Toutanova. 2018.
\newblock Bert: Pre-training of deep bidirectional transformers for language
  understanding.
\newblock \emph{arXiv preprint arXiv:1810.04805}.

\bibitem[{Dielmann and Renals(2008)}]{dielmann2008recognition}
Alfred Dielmann and Steve Renals. 2008.
\newblock Recognition of dialogue acts in multiparty meetings using a switching
  dbn.
\newblock \emph{IEEE transactions on audio, speech, and language processing},
  16(7):1303--1314.

\bibitem[{He et~al.(2016)He, Zhang, Ren, and Sun}]{he2016deep}
Kaiming He, Xiangyu Zhang, Shaoqing Ren, and Jian Sun. 2016.
\newblock Deep residual learning for image recognition.
\newblock In \emph{Proceedings of the IEEE conference on computer vision and
  pattern recognition}, pages 770--778.

\bibitem[{Hochreiter and Schmidhuber(1997)}]{hochreiter1997long}
Sepp Hochreiter and J{\"u}rgen Schmidhuber. 1997.
\newblock Long short-term memory.
\newblock \emph{Neural computation}, 9(8):1735--1780.

\bibitem[{Ji et~al.(2016)Ji, Haffari, and Eisenstein}]{ji2016latent}
Yangfeng Ji, Gholamreza Haffari, and Jacob Eisenstein. 2016.
\newblock A latent variable recurrent neural network for discourse relation
  language models.
\newblock In \emph{Proceedings of NAACL-HLT}, pages 332--342.

\bibitem[{Jurafsky et~al.(1997)Jurafsky, Shriberg, and
  Biasca}]{jurafsky1997switchboard}
Dan Jurafsky, Elizabeth Shriberg, and Debra Biasca. 1997.
\newblock Switchboard swbd-damsl labeling project coder’s manual.
\newblock \emph{Draft 13. Technical Report 97-02}.

\bibitem[{Kalchbrenner and Blunsom(2013)}]{kalchbrenner2013recurrent}
Nal Kalchbrenner and Phil Blunsom. 2013.
\newblock Recurrent convolutional neural networks for discourse
  compositionality.
\newblock In \emph{Proceedings of the Workshop on Continuous Vector Space
  Models and their Compositionality}, pages 119--126.

\bibitem[{Khanpour et~al.(2016)Khanpour, Guntakandla, and
  Nielsen}]{khanpour2016dialogue}
Hamed Khanpour, Nishitha Guntakandla, and Rodney Nielsen. 2016.
\newblock Dialogue act classification in domain-independent conversations using
  a deep recurrent neural network.
\newblock In \emph{Proceedings of COLING 2016, the 26th International
  Conference on Computational Linguistics: Technical Papers}, pages 2012--2021.

\bibitem[{Kim et~al.(2010)Kim, Cavedon, and Baldwin}]{kim2010classifying}
Su~Nam Kim, Lawrence Cavedon, and Timothy Baldwin. 2010.
\newblock Classifying dialogue acts in one-on-one live chats.
\newblock In \emph{Proceedings of the 2010 Conference on Empirical Methods in
  Natural Language Processing}, pages 862--871. Association for Computational
  Linguistics.

\bibitem[{Kumar et~al.(2018)Kumar, Agarwal, Dasgupta, and
  Joshi}]{kumar2018dialogue}
Harshit Kumar, Arvind Agarwal, Riddhiman Dasgupta, and Sachindra Joshi. 2018.
\newblock Dialogue act sequence labeling using hierarchical encoder with crf.
\newblock In \emph{Thirty-Second AAAI Conference on Artificial Intelligence}.

\bibitem[{Lee and Dernoncourt(2016)}]{lee2016sequential}
Ji~Young Lee and Franck Dernoncourt. 2016.
\newblock Sequential short-text classification with recurrent and convolutional
  neural networks.
\newblock In \emph{Proceedings of the 2016 Conference of the North American
  Chapter of the Association for Computational Linguistics: Human Language
  Technologies}, pages 515--520.

\bibitem[{Li et~al.(2017)Li, Su, Shen, Li, Cao, and Niu}]{I17-1099}
Yanran Li, Hui Su, Xiaoyu Shen, Wenjie Li, Ziqiang Cao, and Shuzi Niu. 2017.
\newblock \href {http://aclweb.org/anthology/I17-1099} {Dailydialog: A manually
  labelled multi-turn dialogue dataset}.
\newblock In \emph{Proceedings of the Eighth International Joint Conference on
  Natural Language Processing (Volume 1: Long Papers)}, pages 986--995. Asian
  Federation of Natural Language Processing.

\bibitem[{Liu et~al.(2017)Liu, Han, Tan, and Lei}]{liu2017using}
Yang Liu, Kun Han, Zhao Tan, and Yun Lei. 2017.
\newblock Using context information for dialog act classification in dnn
  framework.
\newblock In \emph{Proceedings of the 2017 Conference on Empirical Methods in
  Natural Language Processing}, pages 2170--2178.

\bibitem[{Milajevs and Purver(2014)}]{milajevs2014investigating}
Dmitrijs Milajevs and Matthew Purver. 2014.
\newblock Investigating the contribution of distributional semantic information
  for dialogue act classification.
\newblock In \emph{Proceedings of the 2nd Workshop on Continuous Vector Space
  Models and their Compositionality (CVSC)}, pages 40--47.

\bibitem[{Nakanishi et~al.(2018)Nakanishi, Inoue, Nakamura, Takanashi, and
  Kawahara}]{nakanishi2018generating}
Ryosuke Nakanishi, Koji Inoue, Shizuka Nakamura, Katsuya Takanashi, and Tatsuya
  Kawahara. 2018.
\newblock Generating fillers based on dialog act pairs for smooth turn-taking
  by humanoid robot.
\newblock In \emph{Proc. Int’l Workshop Spoken Dialogue Systems (IWSDS)}.

\bibitem[{Pennington et~al.(2014)Pennington, Socher, and
  Manning}]{pennington2014glove}
Jeffrey Pennington, Richard Socher, and Christopher Manning. 2014.
\newblock Glove: Global vectors for word representation.
\newblock In \emph{Proceedings of the 2014 conference on empirical methods in
  natural language processing (EMNLP)}, pages 1532--1543.

\bibitem[{Peters et~al.(2018)Peters, Neumann, Iyyer, Gardner, Clark, Lee, and
  Zettlemoyer}]{peters2018deep}
Matthew Peters, Mark Neumann, Mohit Iyyer, Matt Gardner, Christopher Clark,
  Kenton Lee, and Luke Zettlemoyer. 2018.
\newblock Deep contextualized word representations.
\newblock In \emph{Proceedings of the 2018 Conference of the North American
  Chapter of the Association for Computational Linguistics: Human Language
  Technologies, Volume 1 (Long Papers)}, pages 2227--2237.

\bibitem[{Raheja and Tetreault(2019)}]{raheja2019dialogue}
Vipul Raheja and Joel Tetreault. 2019.
\newblock Dialogue act classification with context-aware self-attention.
\newblock \emph{arXiv preprint arXiv:1904.02594}.

\bibitem[{Searle(1969)}]{searle1969speech}
John~Rogers Searle. 1969.
\newblock \emph{Speech acts: An essay in the philosophy of language}, volume
  626.
\newblock Cambridge university press.

\bibitem[{Shaw et~al.(2018)Shaw, Uszkoreit, and Vaswani}]{shaw2018self}
Peter Shaw, Jakob Uszkoreit, and Ashish Vaswani. 2018.
\newblock Self-attention with relative position representations.
\newblock In \emph{Proceedings of the 2018 Conference of the North American
  Chapter of the Association for Computational Linguistics: Human Language
  Technologies, Volume 2 (Short Papers)}, volume~2, pages 464--468.

\bibitem[{Shi and Huang(2018)}]{shi2018dp}
Zhouxing Shi and Minlie Huang. 2018.
\newblock A deep sequential model for discourse parsing on multi-party
  dialogues.
\newblock \emph{arXiv preprint arXiv:1812.00176}.

\bibitem[{Stolcke et~al.(2000)Stolcke, Ries, Coccaro, Shriberg, Bates,
  Jurafsky, Taylor, Martin, Ess-Dykema, and Meteer}]{stolcke2000dialogue}
Andreas Stolcke, Klaus Ries, Noah Coccaro, Elizabeth Shriberg, Rebecca Bates,
  Daniel Jurafsky, Paul Taylor, Rachel Martin, Carol~Van Ess-Dykema, and Marie
  Meteer. 2000.
\newblock Dialogue act modeling for automatic tagging and recognition of
  conversational speech.
\newblock \emph{Computational linguistics}, 26(3):339--373.

\bibitem[{Tavafi et~al.(2013)Tavafi, Mehdad, Joty, Carenini, and
  Ng}]{tavafi2013dialogue}
Maryam Tavafi, Yashar Mehdad, Shafiq Joty, Giuseppe Carenini, and Raymond Ng.
  2013.
\newblock Dialogue act recognition in synchronous and asynchronous
  conversations.
\newblock In \emph{Proceedings of the SIGDIAL 2013 Conference}, pages 117--121.

\bibitem[{Vaswani et~al.(2017)Vaswani, Shazeer, Parmar, Uszkoreit, Jones,
  Gomez, Kaiser, and Polosukhin}]{vaswani2017attention}
Ashish Vaswani, Noam Shazeer, Niki Parmar, Jakob Uszkoreit, Llion Jones,
  Aidan~N Gomez, {\L}ukasz Kaiser, and Illia Polosukhin. 2017.
\newblock Attention is all you need.
\newblock In \emph{Advances in Neural Information Processing Systems}, pages
  5998--6008.

\bibitem[{Wang et~al.(2017)Wang, Li, and Wang}]{wang2017dp}
Yizhong Wang, Sujian Li, and Houfeng Wang. 2017.
\newblock A two-stage parsing method for text-level discourse analysis.
\newblock In \emph{Proceedings of the 55th Annual Meeting of the Association
  for Computational Linguistics (Volume 2: Short Papers)}, volume~2, pages
  184--188.

\bibitem[{Wu et~al.(2016)Wu, Schuster, Chen, Le, Norouzi, Macherey, Krikun,
  Cao, Gao, Macherey et~al.}]{wu2016google}
Yonghui Wu, Mike Schuster, Zhifeng Chen, Quoc~V Le, Mohammad Norouzi, Wolfgang
  Macherey, Maxim Krikun, Yuan Cao, Qin Gao, Klaus Macherey, et~al. 2016.
\newblock Google's neural machine translation system: Bridging the gap between
  human and machine translation.
\newblock \emph{arXiv preprint arXiv:1609.08144}.

\bibitem[{Yang et~al.(2018)Yang, Tu, Wong, Meng, Chao, and
  Zhang}]{yang2018modeling}
Baosong Yang, Zhaopeng Tu, Derek~F Wong, Fandong Meng, Lidia~S Chao, and Tong
  Zhang. 2018.
\newblock Modeling localness for self-attention networks.
\newblock In \emph{Proceedings of the 2018 Conference on Empirical Methods in
  Natural Language Processing}, pages 4449--4458.

\bibitem[{Zhao et~al.(2018)Zhao, Lee, and Eskenazi}]{zhao2018dg}
Tiancheng Zhao, Kyusong Lee, and Maxine Eskenazi. 2018.
\newblock \href {http://aclweb.org/anthology/P18-1101} {Unsupervised discrete
  sentence representation learning for interpretable neural dialog generation}.
\newblock In \emph{Proceedings of the 56th Annual Meeting of the Association
  for Computational Linguistics (Volume 1: Long Papers)}, pages 1098--1107.
  Association for Computational Linguistics.

\bibitem[{Zhao et~al.(2017)Zhao, Zhao, and Eskenazi}]{zhao2017learning}
Tiancheng Zhao, Ran Zhao, and Maxine Eskenazi. 2017.
\newblock Learning discourse-level diversity for neural dialog models using
  conditional variational autoencoders.
\newblock \emph{arXiv preprint arXiv:1703.10960}.

\end{thebibliography}
\bibliographystyle{acl_natbib}

\end{document}